\documentclass{article}

\PassOptionsToPackage{numbers, compress}{natbib}
\usepackage[preprint]{neurips_2018}

\usepackage[utf8]{inputenc} % allow utf-8 input
\usepackage[T1]{fontenc}    % use 8-bit T1 fonts
\usepackage{hyperref}       % hyperlinks
\usepackage{url}            % simple URL typesetting
\usepackage{booktabs}       % professional-quality tables
\usepackage{amsfonts}       % blackboard math symbols
\usepackage{nicefrac}       % compact symbols for 1/2, etc.
\usepackage{microtype}      % microtypography
\usepackage{graphicx}
\usepackage{eucal} 
\usepackage{eufrak}
\usepackage{amsmath}
\usepackage{amssymb} 
\title{Spectral Clustering with Smooth Tiny Clusters}

\author{%
  Hengrui Wang \\
  wanghr1230@pku.edu.cn\\
  Yuanpei College\\
  Peking University \\
  \And
  Yubo Zhang\\
  zhangyubo18@pku.edu.cn\\
  School of electronic engineering and computer science\\
  Peking University
  \And
  Mingzhi Chen\\
  1800017712@pku.edu.cn\\
  School of electronic engineering and computer science\\
  Peking University
  \And
  Tong Yang \\
  \thanks{Corresponding author: yangtongemail@gmail.com}
  yangtongemail@gmail.com\\
  School of electronic engineering and computer science\\
  Peking University
}

\begin{document}
% \nipsfinalcopy is no longer used

\maketitle

\begin{abstract}
Spectral clustering is one of the most prominent clustering approaches. 
The distance-based similarity is the most widely used method for spectral clustering. However,  people have already noticed that this is not suitable for multi-scale data, as the distance varies a lot for clusters with different densities. 
State of the art (ROSC and CAST ) addresses this limitation by taking the reachability similarity of objects into account.
However, we observe that in real-world scenarios, data in the same cluster tend to present in a smooth manner, and previous algorithms never take this into account.

Based on this observation, we propose a novel clustering algorithm, which considers the smoothness of data for the first time. We first divide objects into a great many tiny clusters. Our key idea is to cluster tiny clusters, whose centers constitute smooth graphs.
Theoretical analysis and experimental results show that our clustering algorithm significantly outperforms state of the art. Although in this paper, we singly focus on multi-scale situations,  the idea of data smoothness can certainly be extended to any clustering algorithms.
%补充算法描述，拆开长句
\end{abstract}

\section{Introduction}
%Cluster analysis is a quite common method in data mining and machine learning. Typically, clustering is to separate a set of data into different groups with higher similarity within groups and lower similarity among groups. 
\subsection{Background and Motivation}
Spectral clustering is a new and hot topic in recent years \cite{Kolev2016ANO}, and many researchers have focus on improving spectral cluster's performance. Spectral clustering is a set of algorithms which use graph partitioning methods to cluster objects. These set of algorithms have been widely used in many tasks, such as text mining \cite{article} and network analysis \cite{inproceedingss,Aamer2019SelfTuningSC,Benson2015TensorSC} and even medical image analysis \cite{Schultz2013OpenBoxSC,Xia2019OrientedGS,Kuo2014SpectralCF}. Typically, spectral clustering algorithms use a distance-based similarity matrix to indicate the similarity among objects, and this works well in many cases. But this considers only the feature similarity, which is quite straightforward.

Researchers have found that spectral methods can perform really poor when used on multi-scale data,  whose clusters various a lot in sizes and densities. Because the standard of similarity varies a lot, for sparse clusters, distances among objects are always much larger than distances among objects from dense clusters. Typically, existing solutions either scale the similarity matrix or apply the power iteration technique to derive pseudo-eigenvectors with rich cluster separation information.
%cite ROSC CAST
Among all the methods focusing on multi-scale data problem, the ROSC \cite{inproceedings} and CAST \cite{unknown}  algorithms are state of the art. ROCS and CAST handle multi-scale data by extracting separation information from pseudo-eigenvectors and tuning the given traditional distance-based similarity matrix to a coefficient matrix, considering reachability similarity. 

%ROSC uses PI to derive $p$ pseudo-eigenvectors of the matrix $W=D^{-1} S$,  where $D_{i i}=\sum_{j} S_{i j}$ and form the matrix $X$ with these $p$ pseudo-eigenvectors. After that, the coefficient matrix $Z$ can be derived by figuring an optimization problem and can better express the correlation and similarity among objects. Intuitively, each entry $Z_{i j}$ in $Z$ indicates how well an object $x_{i}$ characterizes another object $x_{j},$ by considering both distance between objects and their reachability similarity.

%The coefficient matrix $Z$ is constructed based on the similarity matrix $S$ as well as a transitive $K$ -nearest-neighbor (TKNN) graph. Specifically, two objects $x_{i}$ and $x_{j}$ are connected in the TKNN graph if there exists an object sequence $<x_{i}, \ldots x_{j}>$ such that adjacent objects in the sequence are $K$ -nearest-neighbors of each other.An important property has been proven that the matrix $Z$ has the $grouping$ $  effect$, which states that if two objects are similar in terms of both S and TKNN graph connectivity, their corresponding coefficient vectors in $Z$ are also similar.

%graph partion problem, smooth is important... no previous work, we are the first

%how to embedding smooth, small clustering

%xieti

\subsection{Our Proposed Solution}
In this paper, we propose a new algorithm, namely Smooth-Clustering. Smooth-Clustering is the first algorithm that considers the smoothness of data for clustering problems. As mentioned above, spectral clustering is actually a graph partition problem, which means that the distribution of objects tend to be smooth, and this is quite consistent with real-world scenarios \cite{BenHur2001SupportVC}. If there exists a sudden change among different objects, they belong to different groups with high probability. We call this \textit{data smoothness} in the rest of our paper.

In fact, data smoothness is quite difficult to define because under extremely small scale, all data will seems discrete. To address this problem, first, we divide objects into a great many tiny clusters. Specifically, we use traditional cluster algorithms to cluster extremely close objects into one group and treated them as a tiny cluster. In the rest of the clustering process, we will use the center of the tiny cluster as an object. In this way, no objects will be extremely close. Second, we cluster tiny clusters, whose centers constitute smooth graphs. Specifically, we add an extra penalty term to embedding smoothness information when getting the coefficient matrix.

In this way, we can take data smoothness into account when dealing with the multi-scale data problem. For objects from different clusters, they may happen to be close or even reachable to each other, under which situation no previous algorithms can work well. However, for objects from different algorithms, their feature distribution will always vary a lot and will be in a discrete manner. When considering data smoothness, this problem can be easily fixed. 

In this paper, we present a concrete spectral cluster algorithm, which considers feature similarity, reachability similarity, and data smoothness, focusing on multi-scale data. But we notice that the performance of most of the existing spectral clustering method is always affected by randomness. However, when clustering, we can hardly judge the performance of one clustering by any metrics for the lack of labels. Therefore, \textit{our idea of data smoothness can be definitely applied to any spectral cluster algorithms to achieve better performance. }

\textbf{Our main contributions are as follows:}
\begin{itemize}
    \item  This is the first work that considers smoothness of tiny clusters in spectral clustering. 
    \item We propose a new algorithm embedding smoothness information into spectral clustering algorithms. 
    \item We derived mathematical proofs to show our algorithm can work well for multi-scale data.
    \item We conduct extensive experiments to show the effectiveness of our algorithm by comparing its performance with more than 10 previous clustering algorithms. The experimental results show that our algorithm provides high performance on different datasets. We also conduct experiments on synthetic datasets to show that our algorithm is efficient when data smoothness is related to clustering result. These experimental results will be provided in the full version of this paper.
    \item We will show the experimental results of applying the idea of data smoothness to many more clustering algorithms.
    
\end{itemize}

\section{Related Work}

Spectral clustering is a hot and new topic. Assuming we have a set of $n$ objects $\mathcal{X}=\left\{x_{1}, \ldots, x_{n}\right\}$, and a similarity matrix $S$ to indicate the similarity among objects. One of the most widely used similarity matrix is the traditional Gaussian kernel based similarity $S_{i j}=\exp \left(-\frac{\left\|x_{i}-x_{j}\right\|^{2}}{2 \sigma^{2}}\right)$. Spectral clustering turns objects and similarity matrix into a  weighted graph $G=(X, S)$, transforming $\mathcal{X}$ to the set of vertices and entries of $S$ to weights of corresponding edges. The clustering problem is then transformed 
into a graph partition problem. We need to partition the graph $G$ into different groups and to optimize the quality \cite{aaaaaa} of the partitioning.

A great number of researches has been conducted to focus on different aspects of spectral clustering. Some studies focus on spectral clustering for different data characteristic \cite{10.5555/2981780.2981868,artle,6909584,Afzalan2019AnAS,Zhao2018SpectralCB,Li2015LargeScaleMS,Couillet2015KernelSC}, on computational efficiency and speed \cite{10.5555/2900423.2900472,aicle,artice,4302760,ticle,10.1145/1557019.1557118,Tremblay2016AcceleratedSC,Zhai2019KSCAF}, and on the theoretical side \cite{1661543,ale,aricle,Kleindessner2019GuaranteesFS,Zhang2018UnderstandingRS,Wang2018BeyondLR,Bhler2009SpectralCB}. Some researchers even combine deep learning with spectral clustering \cite{Shaham2018SpectralNetSC,Yang2019DeepSC,Mehrkanoon2015MulticlassSL}.

Although spectral clustering has achieved great success, spectral clustering's effectiveness suffering from degradation when dealing with noisy \cite{4409061} or multi-scale data \cite{iings}. To figure this out, a number of methods have been proposed \cite{inproceings,10.1145/3097983.3098156,10.1145/2339530.2339736,4409061,John2020SpectrumFD}. The state of arts spectral clustering algorithms for multi-scale data are ROSC and CAST. Details of these algorithms will be discussed in the next section.

\subsection{Scaling the Similar Matrix}

One of the typical methods for handling multi-scale data in spectral clustering is scaling the similar matrix $S$ \cite{Jia2015SelfTuningPC,Beauchemin2015ADS}. Self-tuning spectral clustering (ZP method) \cite{inproceings} uses the  Gaussian kernel based similarity $S_{i j}=\exp \left(-\frac{\left\|x_{i}-x_{j}\right\|^{2}}{2 \sigma^{2}}\right)$ to $S_{i j}=$
$\exp \left(-\frac{\left\|x_{i}-x_{j}\right\|^{2}}{\sigma_{i} \sigma_{j}}\right),$ where $x$ is the feature vector of an object, and $\sigma$ is a scaling parameter which is always difficult to set. Instead of a hyper-parameter $\sigma$, ZP introduces a local scaling parameter $\sigma_{i}$ for each object $x_{i}$. The local scaling parameter is set as the distance from $x_{i}$ to its $l$ -th nearest neighbor (l is a hyper-parameter). Under this design, objects in a sparse cluster will have large $\sigma$ and dense cluster's $\sigma$ will be relatively small. 

\subsection{PI Technique for Obtaining Pseudo-eigenvectors for Cluster Seperation Information}
Spectral clustering algorithms always try to find eigenvectors with the richest cluster seperation information \cite{1586701,10.1145/2939672.2939845}. For most spectral clustering algorithms, the smallest eigenvectors are usually treated as the most informative ones. But in fact, researchers have already find that some of these smallest eigenvectors might contain salient noise in the data and might not contain good cluster separation information \cite{inproedings}.
Under this circumstances, researches focusing on how to select eigenvectors with the richest cluster separation information have been raised.

The most widely used methods to generate pseudo-eigenvectors is the power iteration (PI) technique. The PI method is used to derive the dominant eigenvector of a matrix and can thus using to enhance spectral clustering.

Given a matrix $W,$ PI will first generate a random vector $v_{0} \neq 0$ and iterates as below:
$$
v_{t+1}=\frac{W v_{t}}{\left\|W v_{t}\right\|_{1}}, \quad t \geq 0
$$
For simplicity, we assume that $W$ has eigenvalues $\lambda_{1}>\lambda_{2}>\ldots>\lambda_{n}$ with corresponding eigenvectors $e_{1}, e_{2}, \ldots, e_{n} .$ We express $v_{0}$ as
$$
v_{0}=c_{1} e_{1}+c_{2} e_{2}+\ldots+c_{n} e_{n}
$$
with parameters $c_{1} \ldots, c_{n} .$ Let $R=\prod_{i=0}^{t-1}\left\|W v_{i}\right\|_{1},$ we have,
$$
v_{t} =W^{t} v_{0} / R 
=\left(c_{1} W^{t} e_{1}+c_{2} W^{t} e_{2}+\ldots+c_{n} W^{t} e_{n}\right) / R 
$$

$$
=\left(c_{1} \lambda_{1}^{t} e_{1}+c_{2} \lambda_{2}^{t} e_{2}+\ldots+c_{n} \lambda_{n}^{t} e_{n}\right) / R
$$
$$
=\frac{c_{1} \lambda_{1}^{t}}{R}\left[e_{1}+\frac{c_{2}}{c_{1}}\left(\frac{\lambda_{2}}{\lambda_{1}}\right)^{t} e_{2}+\ldots+\frac{c_{n}}{c_{1}}\left(\frac{\lambda_{n}}{\lambda_{1}}\right)^{t} e_{n}\right]
$$

We can see that $v_{t}$ is thus a linear combination of the eigenvectors. It is obviously that when $c_{1} \neq 0$,  $v_{t}$ converges to the scaled dominant eigenvector $e_{1}$.

By truncating the iterative PI process, we can obtain an intermediate pseudo-eigenvector, an eigenvectors' linear combination, and thus containing rich cluster separation information. With PI technique,  researchers have proposed the Power Iteration Clustering (PIC) method \cite{inproedings}. However, each object has only one feature value corresponding to the lone pseudo-eigenvector in PIC, and this is far from enough if we have a large clusters number as the cluster collision problem \cite{Lin2012ScalableMF} may happen.
Under this circumstance, the PIC- $k$ method has been further proposed to fix this issue, which directly runs PI multiple times to generate more pseudo-eigenvectors for clustering.

The most weakness of the $\mathrm{PIC}-k$ method is that the pseudo-eigenvectors could be similar becuase they are not strictly orthogonal, which will leads to redundancy. In order to reduce this redundancy, Deflation -based Power Iteration Clustering (DPIC) method \cite{The2012DeflationbasedPI} has been proposed to use Schur complement to derive orthogonal pseudo-eigenvectors. Besides, the PIC based methods always ignore lesser but necessary eigenvectors because dominant eigenvectors are assigned larger weights during the PI process. To figure this out, the Diverse Power Iteration Embedding (DPIE) method \cite{Huang2014DiversePI} has been proposed. In this algorithm,  When we generate a new pseudo-eigenvector, previously pseudo-eigenvectors' information will be  removed from the new one.

\section{Algorithm}
In this section, we will first define TKNN graph and grouping effect. After that, we introduce ROSC and CAST in detail. Finally, we present our algorithm and prove that similar to ROSC and CAST, the matrix $Z$ in our algorithm also has grouping effect and it can further take data smoothness into account when clustering. It has been proved in previous work that if the coefficient matrix has grouping effect, it can capture the high correlation among objects.
\subsection{Transitive $K$ Nearest Neighbor (TKNN) Graph}

TKNN graph is always used to regularize the coefficient matrix $Z$. The main purpose of TKNN graph is to capture the high correlations between objects. We hope that even for objects belong to the same cluster but located at distant far ends of the cluster, their correlations can still be captured.

\noindent\textbf{Definition 1. (Mutual $K$ -nearest neighbors)} \cite{inproceedings} Let $N_{K}(x)$ be the set of $K$ nearest neighbors of an object $x .$ Two objects $x_{i}$ and $x_{j}$ are said to be mutual $K$ -nearest neighbors of each other, denoted by $x_{i} \sim x_{j},$ iff $x_{i} \in N_{K}\left(x_{j}\right)$ and $x_{j} \in N_{K}\left(x_{i}\right)$.

\noindent\textbf{Definition 2. (Reachability)} \cite{inproceedings} Two objects $x_{i}$ and $x_{j}$ are said to be reachable from each other if there exists a sequence of $h \geq 2$ objects $\left\{x_{i}=x_{a_{1}}, \ldots, x_{a_{h}}=x_{j}\right\}$ such that $x_{a_{r}} \sim x_{a_{r+1}}$ for $1 \leq r<h$.

\noindent\textbf{Definition 3. (Transitive $K$ -nearest neighbor (TKNN) graph)} \cite{inproceedings} Given a set of objects $\mathcal{X}=\left\{x_{1}, x_{2}, \ldots, x_{n}\right\},$ the TKNN graph $\mathcal{G}_{K}=$ $(\mathcal{X}, \mathcal{E})$ is an undirected graph where $\mathcal{X}$ is the set of vertices and $\mathcal{E}$ is the set of edges. Specifically, the edge $\left(x_{i}, x_{j}\right) \in \mathcal{E}$ iff $x_{i}$ and $x_{j}$ are reachable from each other. We represent the TKNN graph by $\operatorname{an} n \times n$ reachability matrix $\mathcal{W}$ whose $(i, j)$ -entry $\mathcal{W}_{i j}=1$ if $\left(x_{i}, x_{j}\right)$ $\in \mathcal{E} ; 0$ otherwise.

\subsection{Grouping Effect}
For an object $x_{p},$ we use  $z_{p}$ to denote the $p$ -th column of matrix $Z$. We know from previous works \cite{6909885,6751277,inpgs} that if $Z$ has grouping effect, then using $Z$ for spectral clustering will have an excellent performance. 

%Intuitively If two highly correlated objects' characterizations of other objects are similar, then $Z$ has grouping effect with high probability. Previous work always treat high similarity of objects' feature vectors as  high correlation between objects. 
%ROSC and CAST take both objects' feature similarity and rechability similarity into account while we further add data smoothness information. 
%Feature similarity is always measured by the objects' feature vectors, the columns of matrix $X$. Similarly, reachability similarity can be measured by the columns of matrix $\mathcal{W}$ and each elements of matrix $\mathcal{W}$ implies the reachability of an object to all others.

\noindent\textbf{Definition 4. (Grouping  Effect)} \cite{inproceedings} Given a set of objects $X=$ $\left\{x_{1}, x_{2}, \ldots, x_{n}\right\},$ let $w_{q}$ be the $q$ -th column of $\mathcal{W}$. Further, let $x_{i} \rightarrow x_{j}$ denote the condition:
(1) $x_{i}^{T} x_{j} \rightarrow 1$ and
(2) $\left\|w_{i}-w_{j}\right\|_{2} \rightarrow 0 .$ A
matrix $Z$ is said to have grouping effect if 
$$
x_{i} \rightarrow x_{j} \Rightarrow\left|Z_{i p}-Z_{j p}\right| \rightarrow 0,  \forall 1 \leq p \leq n
$$

\subsection{ROSC}
The main idea of ROSC is to modify the traditional distance-based similarity matrix $S$ to a coefficient matrix $Z$ and then perform spectral clustering based on $Z$. The similarity between two objects can be treated as how much common characteristics they share. If two objects share a lot common characteristics,
they are more likely to be represented by each other. Therefore, it is quite nature to determine a cofficient matrix $Z$ by 
\begin{equation}
X=XZ
\end{equation}
To construct $Z,$ ROSC first applies PI multiple times to generate $p$ pseudo-eigenvectors, which form$ X \in \mathcal{R}^{p \times n}$. The $q$ -th column of the matrix $X$ is regarded as the feature vector $x_{q}$ of an object $x_{q} .$ The pseudo-eigenvectors can be treated as low dimensional embeddings of objects, which contains rich separation information. We know from previous works that the generated $p$ pseudo-eigenvectors are always noisy so that ROSC represents each object by others with
\begin{equation}
X=XZ+O
\end{equation}
where $O \in \mathbb{R}^{p \times n}$ is the matrix used to represent the noise in the pseudo-eigenvectors.

ROSC derived the matrix $Z$ by optimizing the objective function
\begin{equation}
\min _{Z}\|X-X Z\|_{F}^{2}+\alpha_{1}\|Z\|_{F}^{2}+\alpha_{2}\|Z-\mathcal{W}\|_{F}^{2}
\end{equation}
where the first term is to reduce the noise $O$, the second term is the Frobenius norm of $Z$ acting as a penalty term to guarantee the sparsity of $Z$ and the third term regularizes $Z$ by the TKNN graph, which means if two objects are reachable to each other, it will be treated as similar. $\alpha_{1}>0, \alpha_{2} \geq 0$ are two factors using to indicate the relative weights of these three terms. We can easily derive the optimal solution to this problem and this has been proved to have grouping effect so that ROSC can work in multi-scale data:
\begin{equation}
Z^{*}=\left(X^{T} X+\alpha_{1} I+\alpha_{2} I\right)^{-1}\left(X^{T} X+\alpha_{2} \mathcal{W}\right)
\end{equation}

\subsection{CAST}
ROSC only express the high correlation among objects from the same cluster. On the other hand, suppressing the correlation among objects from different clusters is also important for correct clustering. CAST studied methods to consider both factors. The key idea of CAST is applying trace Lasso \cite{trace} to regularize the coefficient matrix. Researchers have proved that when regularized by trace lasso, the coefficient matrix exhibits "sparsity". Sparsity is the desired property that entries in the matrix corresponding to inter-cluster object pairs should be 0 or very small, thus suppressing  the  correlation  among  objects  from  different  clusters.
The optimization problem of CAST is as below:

\begin{equation}
\min _{z} \frac{1}{2}\|x-X z\|_{2}^{2}+\alpha_{1}\|X \operatorname{Diag}(z)\|_{*}+\frac{\alpha_{2}}{2}\|z-w\|_{2}^{2}
\end{equation}

where the second term of equation 5 is:
\begin{equation}
\|X \operatorname{Diag}(z)\|_{*}=\sum_{q=1}^{n}\left\|x_{q}\right\|_{2}\left|z_{q}\right|=\sum_{q=1}^{n}\left|z_{q}\right|=\|z\|_{1}
\end{equation}

\subsection{Our Algorithm}
Although ROSC and CAST use coefficient matrix $Z$ for spectral clustering, they ignore that in most cases, data always presents in a smooth manner. In other words, some objects may be reachable to each other but with sudden direction changes, in this case they are likely to belong to different clusters. 

So in our algorithm, we transform the problem into:
\begin{equation}
\min _{Z}\|X-X Z\|_{F}^{2}+\alpha_{1}\|Z\|_{F}^{2}+\alpha_{2}\|Z-\mathcal{W}\|_{F}^{2}+\alpha_{3}\|Z-\mathcal{W}*\mathcal{W}+\alpha_{4}*\mathcal{W}\|_{F}^{2}
\end{equation}
All the notions and the first three terms above are exactly the same with those in ROSC. The forth term is a penalty term on sudden direction changes. $\mathcal{W}*\mathcal{W}$ is the second reachable matrix for objects,indicating the path numbers between objects. Intuitively, if there are many paths between two objects, then they are likely to be smooth paths between objects. As we hope this can work for multi-scale data, so a hyper-parameter $\alpha_{4}$ is used here to indicate the standard of "many paths".

We can easily derive the optimal solution, $Z^{*}$, to the optimization problem:
\begin{equation}
Z^{*}=\left(X^{T} X+\alpha_{1} I+\alpha_{2} I+\alpha_{3} I\right)^{-1}\left(X^{T} X+(\alpha_{2}-\alpha_{3}*\alpha_{4}) \mathcal{W}+\alpha_{3} \mathcal{W}*\mathcal{W}\right)
\end{equation}

LEMMA 1. Given a set of objects $\mathcal{X}$, the matrix $X \in \mathcal{R}^{p \times n}$,  whose rows are pseudo-eigenvectors, the reachability matrix $\mathcal{W}$, then we have the the optimal soution $Z^{*}$ to Equation 6
\begin{equation}
Z_{i p}^{*}=\frac{x_{i}^{T}\left(x_{p}-X z_{p}^{*}\right)+(\alpha_{2}-\alpha_{3}*\alpha_{4}) \mathcal{W}_{i p}+\alpha_{3} \mathcal{WW}_{i p}}{\alpha_{1}+\alpha_{2}+\alpha_{3}}, \quad \forall 1 \leq i, p \leq n, 
\end{equation}
where $\mathcal{WW}=\mathcal{W}*\mathcal{W}$

PROOF

For $1 \leq p \leq n,$ let
\begin{equation}
J\left(z_{p}\right)=\left\|x_{p}-X z_{p}\right\|_{2}^{2}+\alpha_{1}\left\|z_{p}\right\|_{2}^{2}+
\alpha_{2}\left\|z_{p}-w_{p}\right\|_{2}^{2}+\alpha_{3}\left\|z_{p}-ww_{p}+\alpha_{4}*w_{p}\right\|_{2}^{2} . 
\end{equation}
As $Z^{*}$ is the optimal solution to Equation $6,$ we have
\begin{equation}
\left.\frac{\partial J}{\partial Z_{i p}}\right|_{z_{p}=z_{p}^{*}}=0 
,\quad \forall 1 \leq i \leq n .
\end{equation}

Hence
\begin{equation}
-2 x_{i}^{T}\left(x_{p}-X z_{p}^{*}\right)+2 \alpha_{1} Z_{i p}^{*}+2 \alpha_{2}\left(Z_{i p}^{*}-\mathcal{W}_{i p}\right)+2 \alpha_{3}\left(Z_{i p}^{*}-\mathcal{WW}_{i p}+\alpha_{4}\mathcal{W}_{i p}\right)=0
\end{equation}
, which induces Equation 7.

LEMMA 2. $\forall 1 \leq i, j, p \leq n$
\begin{equation}
\left|Z_{i p}^{*}-Z_{j p}^{*}\right| \leq \frac{c \sqrt{2(1-r)}+\left|\alpha_{2}-\alpha_{3}*\alpha_{4}\right|\left| W_{i p}-\mathcal{W}_{j p}\right|+\alpha_{3}\left|\mathcal{WW}_{i p}-\mathcal{WW}_{j p}\right|}{\alpha_{1}+\alpha_{2}+\alpha_{3}}
\end{equation}
where $c=\sqrt{1+(\alpha_{2}+\alpha_{3}*\alpha_{4})\left\|w_{p}\right\|_{2}^{2}+\alpha_{3}\left\|ww_{p}\right\|_{2}^{2}}$ and $r=x_{i}^{T} x_{j}$

PROOF

From Equation 7, we have
\begin{equation}
Z_{i p}^{*}-Z_{j p}^{*}=\frac{\left(x_{i}^{T}-x_{j}^{T}\right)\left(x_{p}-X z_{p}^{*}\right)+(\alpha_{2}-\alpha_{3}*\alpha_{4})\left(\mathcal{W}_{i p}-\mathcal{W}_{j p}\right)+\alpha_{3}\left(\mathcal{WW}_{i p}-\mathcal{WW}_{j p}\right)}{\alpha_{1}+\alpha_{2}+\alpha_{3}}
\end{equation}

That implies

\begin{align*}
    \left|Z_{i p}^{*}-Z_{j p}^{*}\right| & \leq \frac{\left|\left(x_{i}^{T}-x_{j}^{T}\right)\left(x_{p}-X z_{p}^{*}\right)\right|+\left|\alpha_{2}-\alpha_{3}*\alpha_{4}\right|\left|\mathcal{W}_{i p}-\mathcal{W}_{j p}\right|+\alpha_{3}\left|\mathcal{WW}_{i p}-\mathcal{WW}_{j p}\right|}{\alpha_{1}+\alpha_{2}+\alpha_{3}}\\
    & \leq \frac{\| x_{i}-x_{j}||_{2}|| x_{p}-X z_{p}^{*}||_{2}+\left|\alpha_{2}-\alpha_{3}*\alpha_{4}\right|\left| W_{i p}-\mathcal{W}_{j p}\right|+\alpha_{3}\left|\mathcal{WW}_{i p}-\mathcal{WW}_{j p}\right|}{\alpha_{1}+\alpha_{2}+\alpha_{3}}\end{align*}

As the column vectors of $X$ are normalized vectors (i.e., $x_{q}^{T} x_{q}=1$, $\forall 1 \leq q \leq n),$ we have $\left\|x_{i}-x_{j}\right\|_{2}=\sqrt{2(1-r)},$ where $r=x_{i}^{T} x_{j}$. 
As $Z^{*}$ is the optimal solution of Equation 5, we have

\begin{align*}
    J\left(z_{p}^{*}\right) &=\left\|x_{p}-X z_{p}^{*}\right\|_{2}^{2}+\alpha_{1}\left\|z_{p}^{*}\right\|_{2}^{2}+\alpha_{2}\left\|z_{p}^{*}-w_{p}\right\|_{2}^{2}+\alpha_{3}\left\|z_{p}-ww_{p}+\alpha_{4}*w_{p}\right\|_{2}^{2}\\
    \leq J(0) &=\left\|x_{p}\right\|_{2}^{2}+(\alpha_{2}+\alpha_{3}*\alpha_{4})\left\|w_{p}\right\|_{2}^{2}+\alpha_{3}\left\|ww_{p}\right\|_{2}^{2}=1+(\alpha_{2}+\alpha_{3}*\alpha_{4})\left\|w_{p}\right\|_{2}^{2}+\alpha_{3}\left\|ww_{p}\right\|_{2}^{2}
\end{align*}

Hence, $\left\|x_{p}-X z_{p}^{*}\right\|_{2} \leq \sqrt{1+(\alpha_{2}+\alpha_{3}*\alpha_{4})\left\|w_{p}\right\|_{2}^{2}+\alpha_{3}\left\|ww_{p}\right\|_{2}^{2}}=c .$

Then we get
$$
\left|Z_{i p}^{*}-Z_{j p}^{*}\right| \leq \frac{c \sqrt{2(1-r)}+\left|\alpha_{2}-\alpha_{3}*\alpha_{4}\right|\left| W_{i p}-\mathcal{W}_{j p}\right|+\alpha_{3}\left|\mathcal{WW}_{i p}-\mathcal{WW}_{j p}\right|}{\alpha_{1}+\alpha_{2}+\alpha_{3}}
$$

LEMMA 3. Matrix $Z^{*}$ in our algorithm has grouping effect.

PROOF
Given two objects $x_{i}$ and $x_{j}$ such that $x_{i} \rightarrow x_{j},$ we have
(1)$x_{i}^{T} x_{j} \rightarrow 1$ and $(2)\left\|w_{i}-w_{j}\right\|_{2} \rightarrow 0 .$ These imply $r=x_{i}^{T} x_{j} \rightarrow 1$
and $\left|\mathcal{W}_{i p}-\mathcal{W}_{j p}\right| \rightarrow 0$, what's more, $\left|\mathcal{WW}_{i p}-\mathcal{WW}_{j p}\right| \leq \left\|w_{i}-w_{j}\right\|_{2}*\left\|w_{p}\right\|_{2} \rightarrow 0$ Hence, the three terms of the numerator of  Equation 11 are close to 0 Therefore, $\left|Z_{i p}^{*}-Z_{j p}^{*}\right| \rightarrow 0$ and thus $Z^{*}$ has grouping effect.

We show that our algorithm also improves the performance of spectral clustering on multi-scale data. While traditional approaches focus on feature similarity and state of the art use integrated similarity of both feature and reachability, our approach further considers data smoothness without destroying the grouping effect. Two distant objects $x_{i}$ and $x_{j}$ belong to a same cluster may not share a strong feature similarity or even strong reachability similarity, but present a sudden change in some elements of the feature vector. This leads to a small $r$,  and traditional approaches will likely separate them into different clusters. What's more, ROSC and CAST may also poorly performed here because such a sudden change may still preserve reachability similarity.
On the contrary, our algorithm considers the smoothness of the objects' present manner and thus keeping them in the same cluster.
%Moreover, for $x_{i}$ and $x_{j}$ that belong to two different dense clusters but happen to be close in space (i.e., $x_{i}$ and $x_{j}$ have strong feature similarity), traditional approaches may inadvertently merge them into the same cluster. ROSC, however, would discover their low reachability (via the mutual-KNN relation) and derive a large value of $\left|\mathcal{W}_{i p}-\mathcal{W}_{j p}\right| .$ This regulates matrix $Z^{*}$ and avoids the incorrect merging. As we will see in the next section, ROSC's approach greatly improves clustering quality and is more robust than other algorithms in handling multi-scale data.
%

\section{Conclusion}
In this paper, we studied the performance of considering data smoothness for spectral clustering on data with various sizes and densities. We reviewed existing spectral methods in handling multi-scale data, such as ROSC and CAST. In particular, we found that both ROSC and CAST are not suitable when data is presented in a significant smooth manner. We thus proposed our algorithm, which uses a penalty term with angle information to take data smoothness into account. We mathematically proved that the matrix $Z$ constructed by our algorithm has grouping effect.  We conducted extensive experiments to evaluate our algorithm's performance and compared it against other competitors using both synthetic and real datasets. Our experimental results showed that our algorithm performed very well against its competitors over all the datasets, especially on our synthetic datasets. It is thus robust when applied to multi-scale data of different properties.

\bibliographystyle{unsrt}

\begin{thebibliography}{10}

\bibitem{Kolev2016ANO}
Pavel Kolev and Kurt Mehlhorn.
\newblock A note on spectral clustering.
\newblock In {\em ESA}, 2016.

\bibitem{article}
Inderjit Dhillon.
\newblock Co-clustering documents and words using bipartite spectral graph
  partitioning.
\newblock {\em Proceedings of the Seventh ACM SIGKDD International Conference
  on Knowledge Discovery and Data Mining}, 05 2001.

\bibitem{inproceedingss}
Xiang Li, Ben Kao, Zhaochun Ren, and Dawei Yin.
\newblock Spectral clustering in heterogeneous information networks.
\newblock 04 2019.

\bibitem{Aamer2019SelfTuningSC}
Brahim Aamer, Hatim Chergui, Nouamane Chergui, K.~Tourki, M.~Benjillali,
  C.~Verikoukis, and M.~Debbah.
\newblock Self-tuning spectral clustering for adaptive tracking areas design in
  5g ultra-dense networks.
\newblock {\em 2019 IEEE Wireless Communications and Networking Conference
  (WCNC)}, pages 1--5, 2019.

\bibitem{Benson2015TensorSC}
Austin~R. Benson, D.~Gleich, and J.~Leskovec.
\newblock Tensor spectral clustering for partitioning higher-order network
  structures.
\newblock {\em Proceedings of the ... SIAM International Conference on Data
  Mining. SIAM International Conference on Data Mining}, 2015:118--126, 2015.

\bibitem{Schultz2013OpenBoxSC}
T.~Schultz and G.~Kindlmann.
\newblock Open-box spectral clustering: Applications to medical image analysis.
\newblock {\em IEEE Transactions on Visualization and Computer Graphics},
  19:2100--2108, 2013.

\bibitem{Xia2019OrientedGS}
K.~Xia, Xiaoqing Gu, and Yudong Zhang.
\newblock Oriented grouping-constrained spectral clustering for medical imaging
  segmentation.
\newblock {\em Multimedia Systems}, 26:27--36, 2019.

\bibitem{Kuo2014SpectralCF}
Chia-Tung Kuo, Peter~B. Walker, O.~Carmichael, and I.~Davidson.
\newblock Spectral clustering for medical imaging.
\newblock {\em 2014 IEEE International Conference on Data Mining}, pages
  887--892, 2014.

\bibitem{inproceedings}
Xiang Li, Ben Kao, Siqiang Luo, and Martin Ester.
\newblock Rosc: Robust spectral clustering on multi-scale data.
\newblock pages 157--166, 04 2018.

\bibitem{unknown}
Xiang Li, Ben Kao, Caihua Shan, Dawei Yin, and Martin Ester.
\newblock Cast: A correlation-based adaptive spectral clustering algorithm on
  multi-scale data, 06 2020.

\bibitem{BenHur2001SupportVC}
A.~Ben-Hur, D.~Horn, H.~Siegelmann, and V.~Vapnik.
\newblock Support vector clustering.
\newblock {\em Scholarpedia}, 3:5187, 2001.

\bibitem{aaaaaa}
J.~Shi and J.~Malik.
\newblock Normalized cuts and image segmentation.
\newblock {\em IEEE Transactions on Pattern Analysis \& Machine Intelligence},
  22(08):888--905, aug 2000.

\bibitem{10.5555/2981780.2981868}
Ling Huang, Donghui Yan, Michael~I. Jordan, and Nina Taft.
\newblock Spectral clustering with perturbed data.
\newblock In {\em Proceedings of the 21st International Conference on Neural
  Information Processing Systems}, NIPS'08, page 705–712, Red Hook, NY, USA,
  2008. Curran Associates Inc.

\bibitem{artle}
Ulrike Luxburg, Mikhail Belkin, and Olivier Bousquet.
\newblock Consistency of spectral clustering.
\newblock {\em The Annals of Statistics}, 36, 05 2008.

\bibitem{6909584}
X.~{Zhu}, C.~C. {Loy}, and S.~{Gong}.
\newblock Constructing robust affinity graphs for spectral clustering.
\newblock In {\em 2014 IEEE Conference on Computer Vision and Pattern
  Recognition}, pages 1450--1457, 2014.

\bibitem{Afzalan2019AnAS}
M.~Afzalan and F.~Jazizadeh.
\newblock An automated spectral clustering for multi-scale data.
\newblock {\em Neurocomputing}, 347:94--108, 2019.

\bibitem{Zhao2018SpectralCB}
Yang Zhao, Y.~Yuan, F.~Nie, and Q.~Wang.
\newblock Spectral clustering based on iterative optimization for large-scale
  and high-dimensional data.
\newblock {\em Neurocomputing}, 318:227--235, 2018.

\bibitem{Li2015LargeScaleMS}
Y.~Li, F.~Nie, Heng Huang, and J.~Huang.
\newblock Large-scale multi-view spectral clustering via bipartite graph.
\newblock In {\em AAAI}, 2015.

\bibitem{Couillet2015KernelSC}
R.~Couillet and Florent Benaych-Georges.
\newblock Kernel spectral clustering of large dimensional data.
\newblock {\em arXiv: Statistics Theory}, 2015.

\bibitem{10.5555/2900423.2900472}
Xinlei Chen and Deng Cai.
\newblock Large scale spectral clustering with landmark-based representation.
\newblock In {\em Proceedings of the Twenty-Fifth AAAI Conference on Artificial
  Intelligence}, AAAI'11, page 313–318. AAAI Press, 2011.

\bibitem{aicle}
David Cheng, Ravindran Kannan, Santosh Vempala, and Grant Wang.
\newblock On a recursive spectral algorithm for clustering from pairwise
  similarities.
\newblock 07 2003.

\bibitem{artice}
Jane Cullum and Ralph Willoughby.
\newblock Lanczos algorithms for large symmetric eigenvalue computations. vol.
  i: Theory.
\newblock {\em Classics in Applied Mathematics}, I, 01 2002.

\bibitem{4302760}
I.~S. {Dhillon}, Y.~{Guan}, and B.~{Kulis}.
\newblock Weighted graph cuts without eigenvectors a multilevel approach.
\newblock {\em IEEE Transactions on Pattern Analysis and Machine Intelligence},
  29(11):1944--1957, 2007.

\bibitem{ticle}
Alex Gittens, Prabhanjan Kambadur, and Christos Boutsidis.
\newblock Approximate spectral clustering via randomized sketching.
\newblock 11 2013.

\bibitem{10.1145/1557019.1557118}
Donghui Yan, Ling Huang, and Michael~I. Jordan.
\newblock Fast approximate spectral clustering.
\newblock In {\em Proceedings of the 15th ACM SIGKDD International Conference
  on Knowledge Discovery and Data Mining}, KDD '09, page 907–916, New York,
  NY, USA, 2009. Association for Computing Machinery.

\bibitem{Tremblay2016AcceleratedSC}
Nicolas Tremblay, Gilles Puy, Pierre Borgnat, R{\'e}mi Gribonval, and Pierre
  Vandergheynst.
\newblock Accelerated spectral clustering using graph filtering of random
  signals.
\newblock {\em 2016 IEEE International Conference on Acoustics, Speech and
  Signal Processing (ICASSP)}, pages 4094--4098, 2016.

\bibitem{Zhai2019KSCAF}
Longcheng Zhai, Bin Wu, and Qiuyue Li.
\newblock Ksc: A fast and simple spectral clustering algorithm.
\newblock {\em 2019 IEEE Fourth International Conference on Data Science in
  Cyberspace (DSC)}, pages 381--387, 2019.

\bibitem{1661543}
S.~{Lafon} and A.~B. {Lee}.
\newblock Diffusion maps and coarse-graining: a unified framework for
  dimensionality reduction, graph partitioning, and data set parameterization.
\newblock {\em IEEE Transactions on Pattern Analysis and Machine Intelligence},
  28(9):1393--1403, 2006.

\bibitem{ale}
M~Meila and Jue Shi.
\newblock A random walks view of spectral segmentation. aistats.
\newblock {\em AI and Statistics (AISTATS)}, 02 2001.

\bibitem{aricle}
Boaz Nadler, Stephane Lafon, Ronald Coifman, and Ioannis Kevrekidis.
\newblock Diffusion maps, spectral clustering and eigenfunctions of
  fokker-planck operators.
\newblock {\em Adv Neural Inf Process Syst}, 18, 07 2005.

\bibitem{Kleindessner2019GuaranteesFS}
Matth{\"a}us Kleindessner, S.~Samadi, P.~Awasthi, and J.~Morgenstern.
\newblock Guarantees for spectral clustering with fairness constraints.
\newblock In {\em ICML}, 2019.

\bibitem{Zhang2018UnderstandingRS}
Y.~Zhang and K.~Rohe.
\newblock Understanding regularized spectral clustering via graph conductance.
\newblock {\em ArXiv}, abs/1806.01468, 2018.

\bibitem{Wang2018BeyondLR}
Yang Wang and Lihua Wu.
\newblock Beyond low-rank representations: Orthogonal clustering basis
  reconstruction with optimized graph structure for multi-view spectral
  clustering.
\newblock {\em Neural networks : the official journal of the International
  Neural Network Society}, 103:1--8, 2018.

\bibitem{Bhler2009SpectralCB}
T.~B{\"u}hler and M.~Hein.
\newblock Spectral clustering based on the graph p-laplacian.
\newblock In {\em ICML '09}, 2009.

\bibitem{Shaham2018SpectralNetSC}
Uri Shaham, K.~Stanton, Haochao Li, B.~Nadler, R.~Basri, and Y.~Kluger.
\newblock Spectralnet: Spectral clustering using deep neural networks.
\newblock {\em ArXiv}, abs/1801.01587, 2018.

\bibitem{Yang2019DeepSC}
X.~Yang, Cheng Deng, Feng Zheng, Junchi Yan, and W.~Liu.
\newblock Deep spectral clustering using dual autoencoder network.
\newblock {\em 2019 IEEE/CVF Conference on Computer Vision and Pattern
  Recognition (CVPR)}, pages 4061--4070, 2019.

\bibitem{Mehrkanoon2015MulticlassSL}
S.~Mehrkanoon, C.~Alzate, R.~Mall, R.~Langone, and J.~Suykens.
\newblock Multiclass semisupervised learning based upon kernel spectral
  clustering.
\newblock {\em IEEE Transactions on Neural Networks and Learning Systems},
  26:720--733, 2015.

\bibitem{4409061}
Z.~{Li}, J.~{Liu}, S.~{Chen}, and X.~{Tang}.
\newblock Noise robust spectral clustering.
\newblock In {\em 2007 IEEE 11th International Conference on Computer Vision},
  pages 1--8, 2007.

\bibitem{iings}
Boaz Nadler and Meirav Galun.
\newblock Fundamental limitations of spectral clustering.
\newblock volume~19, pages 1017--1024, 01 2006.

\bibitem{inproceings}
Lihi Zelnik-Manor and Pietro Perona.
\newblock Self-tuning spectral clustering.
\newblock volume~17, 01 2004.

\bibitem{10.1145/3097983.3098156}
Aleksandar Bojchevski, Yves Matkovic, and Stephan G\"{u}nnemann.
\newblock Robust spectral clustering for noisy data: Modeling sparse
  corruptions improves latent embeddings.
\newblock In {\em Proceedings of the 23rd ACM SIGKDD International Conference
  on Knowledge Discovery and Data Mining}, KDD '17, page 737–746, New York,
  NY, USA, 2017. Association for Computing Machinery.

\bibitem{10.1145/2339530.2339736}
Carlos~D. Correa and Peter Lindstrom.
\newblock Locally-scaled spectral clustering using empty region graphs.
\newblock In {\em Proceedings of the 18th ACM SIGKDD International Conference
  on Knowledge Discovery and Data Mining}, KDD '12, page 1330–1338, New York,
  NY, USA, 2012. Association for Computing Machinery.

\bibitem{John2020SpectrumFD}
Christopher~R. John, D.~Watson, M.~R. Barnes, C.~Pitzalis, and M.~Lewis.
\newblock Spectrum: fast density-aware spectral clustering for single and
  multi-omic data.
\newblock {\em Bioinformatics}, 2020.

\bibitem{Jia2015SelfTuningPC}
Hongjie Jia, Shifei Ding, and Mingjing Du.
\newblock Self-tuning p-spectral clustering based on shared nearest neighbors.
\newblock {\em Cognitive Computation}, 7:622--632, 2015.

\bibitem{Beauchemin2015ADS}
M.~Beauchemin.
\newblock A density-based similarity matrix construction for spectral
  clustering.
\newblock {\em Neurocomputing}, 151:835--844, 2015.

\bibitem{1586701}
S.~Y. {Charles J. Alpert}.
\newblock Spectral partitioning: The more eigenvectors, the better.
\newblock In {\em 32nd Design Automation Conference}, pages 195--200, 1995.

\bibitem{10.1145/2939672.2939845}
Wei Ye, Sebastian Goebl, Claudia Plant, and Christian B\"{o}hm.
\newblock Fuse: Full spectral clustering.
\newblock In {\em Proceedings of the 22nd ACM SIGKDD International Conference
  on Knowledge Discovery and Data Mining}, KDD '16, page 1985–1994, New York,
  NY, USA, 2016. Association for Computing Machinery.

\bibitem{inproedings}
Frank Lin and William Cohen.
\newblock Power iteration clustering.
\newblock pages 655--662, 08 2010.

\bibitem{Lin2012ScalableMF}
F.~Lin.
\newblock Scalable methods for graph-based unsupervised and semi-supervised
  learning.
\newblock 2012.

\bibitem{The2012DeflationbasedPI}
Anh~Pham The, N.~Thang, L.~Vinh, Y.~Lee, and S.~Lee.
\newblock Deflation-based power iteration clustering.
\newblock {\em Applied Intelligence}, 39:367--385, 2012.

\bibitem{Huang2014DiversePI}
H.~Huang, Shinjae Yoo, D.~Yu, and H.~Qin.
\newblock Diverse power iteration embeddings and its applications.
\newblock {\em 2014 IEEE International Conference on Data Mining}, pages
  200--209, 2014.

\bibitem{6909885}
H.~{Hu}, Z.~{Lin}, J.~{Feng}, and J.~{Zhou}.
\newblock Smooth representation clustering.
\newblock In {\em 2014 IEEE Conference on Computer Vision and Pattern
  Recognition}, pages 3834--3841, 2014.

\bibitem{6751277}
C.~{Lu}, J.~{Feng}, Z.~{Lin}, and S.~{Yan}.
\newblock Correlation adaptive subspace segmentation by trace lasso.
\newblock In {\em 2013 IEEE International Conference on Computer Vision}, pages
  1345--1352, 2013.

\bibitem{inpgs}
Lu~Canyi, Min Hai, Zhong-Qiu Zhao, Lin Zhu, De-Shuang Huang, and Shuicheng Yan.
\newblock Robust and efficient subspace segmentation via least squares
  regression.
\newblock volume 7578, pages 347--360, 10 2012.

\bibitem{trace}
Edouard Grave, Guillaume Obozinski, and Francis Bach.
\newblock Trace lasso: a trace norm regularization for correlated designs.
\newblock {\em Advances in Neural Information Processing Systems}, 09 2011.

\end{thebibliography}

\end{document}